\documentclass[letterpaper, 10 pt, conference]{ieeeconf}  

\IEEEoverridecommandlockouts                              

\usepackage[dvipsnames]{xcolor}

\title{\LARGE \bf
Collective Conditioned Reflex: A Bio-Inspired Fast Emergency Reaction Mechanism for Designing Safe Multi-Robot Systems
}

\author{Bowei He$^{1*}$,  Zhenting Zhao$^{1,2,*}$,  Wenhao Luo$^{3}$, Rui Liu$^{1}$

\thanks{$^{*}$ denotes Bowei and Zhenting contribute equally. $^{1}$ are with The Cognitive Robotics and AI Lab (CRAI), College of Aeronautics and Engineering, Kent State University, Kent, OH 44240, USA.  $^{2}$ is with University of California, Berkeley.$^{3}$ is with Department of Computer Science, College of Computing and Informatics, University of North Carolina at Charlotte.  Dr. Rui Liu is the corresponding author, {\tt\small ruiliu.robotics@gmail.com}.
}%
}

\usepackage{enumerate}
\usepackage{algorithm}
\usepackage{algpseudocode}
\usepackage{graphicx}
\usepackage{booktabs}
\usepackage{array}
\usepackage{caption}
\usepackage{soul}
\begin{document}

\maketitle
\thispagestyle{plain}
\pagestyle{plain}

\begin{abstract}

A multi-robot system (MRS) is a group of coordinated robots designed to cooperate with each other and accomplish given tasks. Due to the uncertainties in operating environments, the system may encounter emergencies, such as unobserved obstacles, moving vehicles, and extreme weather. Animal groups such as bee colonies initiate collective emergency reaction behaviors such as bypassing obstacles and avoiding predators, similar to muscle conditioned reflex which organizes local muscles to avoid hazards in the first response without delaying passage through the brain. Inspired by this, we develop a similar collective conditioned reflex mechanism for multi-robot systems to respond to emergencies.
In this study, Collective Conditioned Reflex (CCR), a bio-inspired emergency reaction mechanism, is developed based on animal collective behavior analysis and multi-agent reinforcement learning (MARL). The algorithm uses a physical model to determine if the robots are experiencing an emergency; then, rewards for robots involved in the emergency are augmented with corresponding heuristic rewards, which evaluate emergency magnitudes and consequences and decide local robots' participation. 
CCR is validated on three typical emergency scenarios: \textit{turbulence, strong wind, and hidden obstacle}. Simulation results demonstrate that CCR improves robot teams' emergency reaction capability with faster reaction speed and safer trajectory adjustment compared with baseline methods.

\end{abstract}

\section{INTRODUCTION}


A Multi-robot system (MRS) is a system consisting of a collection of coordinated robots. Compared with a single robot system, MRS often has a larger action scope and can handle more complex tasks with multiple targets, larger workspaces, and more sophiscated mission procedures. For example, multiple unmanned aerial vehicle (UAVs) and multiple unmanned ground vehicle (UGVs) teams have been successfully applied to disaster relief, explosive ordinance disposal, structural inspection, reconnaissance, cargo freight, hostage search, and many other complex scenarios.

MRS deployments usually rely on real-time human teleoperation. However, the complexity of these tasks presents a heavy workload for the operator, leading to delayed or unsafe guidance for robots. Unexpected emergencies that require rapid adjustments of the MRS inevitably happen during missions, exacerbating the problem. This causes performance degradation and physical system damages to the robot team and can even threaten the safety of personnels involved \cite{khamis2015multi,nagatani2013emergency}. Therefore, to ensure task success and safety, an emergency reaction mechanism is highly desired for real-world MRS deployments. 

\begin{figure}
    \centering
    \includegraphics[width=0.47\textwidth]{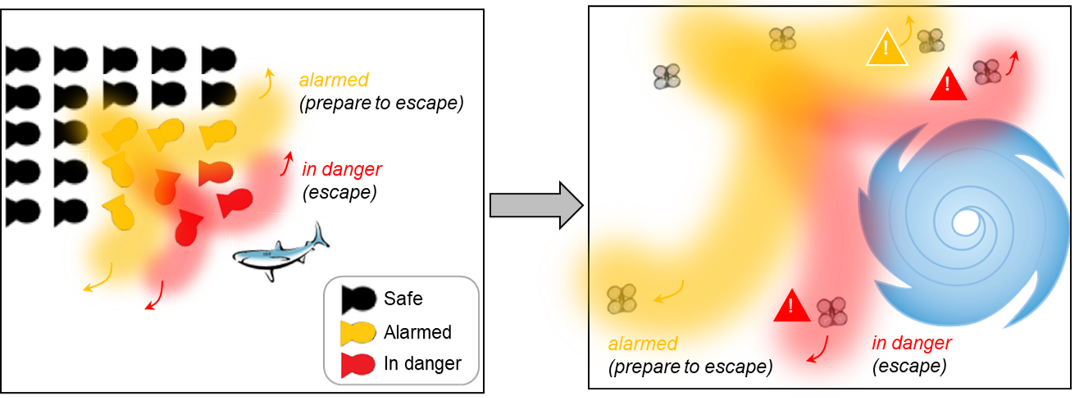}
    \caption{Left: fish schooling for deterring predators. 
    Right: with the proposed CCR, a group of unmanned aerial vehicles (UAV) working collectively to avoid the emergency. 
    }
    \label{fig:fish}
    \vspace{-0.7cm}
\end{figure}

Due to the dynamic and uncertain nature of emergencies and the complexity of MRS control, it is challenging to ensure the safety of MRS in real-world deployments. Traditional autonomous MRS mainly relies on centralized control to plan paths \cite{liu2019trust,pang2020trust}. However, centralized emergency reactions require going through the central decision-making node, limiting reaction speed, and creating a single point of failure which is fragile during an emergency. On the other hand, simple decentralized control suffers from the opposite problem of weak coordination among multiple robots and global suboptimality in task executions. A common drawback of both centralized and decentralized methods is the failure to incorporate the physical properties such as the type and degree of emergent situations, thus decreasing overall task performance. In addition, traditional planning methods lack robustness against unexpected emergencies which have not been sufficiently explored by the robots. 

To address these issues, the bio-inspired Collective Conditioned Reflex (CCR) algorithm is developed for fast emergency reactions. Similar to some other bio-inspired MRS researches~\cite{liu2019trust,pierson2015bio}, CCR is inspired by the collective behaviors of animal groups responding to predators and obstacles \cite{gordon2016division}. When a fish school encounters a predator, the fish making the contact will quickly react and escape together along safe directions. Meanwhile, they also share the emergency information with other nearby fish, who adjust their behaviors to help the whole school to avoid the predator. These behaviors of local-global collective emergency reaction maximally avoid the loss in danger as the most urgent fish/robots react first and reduce response time by using local analysis and avoiding going through a global analysis. This collective intelligence in emergency reaction effectively helps an animal team to avoid danger; the reaction mechanism is similar to spinal reflexes in human muscles, which are fast response mechanisms that react to stimulation without going through the central decision-maker -- the brain \cite{mcgee2021evidence}. These collective behaviors help them react to dangers preemptively and rapidly, providing safety guidance at an early stage for the whole team. Such observations inspire us to apply a similar approach in training multi-robot systems (Fig. \ref{fig:fish}). In particular, we encourage pioneer robots encountering danger to locally collaborate to avoid dangers without going through centralized analysis for global optimal strategy searching; meanwhile, pioneer robots share danger information with other robots to guide their motion behaviors to achieve the goal of team emergency avoidance.

The contributions in this study are mainly three folds:
\begin{enumerate}[(a)]
    \item A novel bio-inspired emergency reaction model is developed by mimicking animals' emergency reaction behavior. The module can be integrated with any existing cooperative multi-agent reinforcement learning algorithm to improve the emergency reaction capabilities of robot teams. 
    \item An environment-adaptive safe learning framework is developed by considering safety assessment and safety adjustment. This framework provides a pipeline for customizing the robot teams' trade-offs between safety and efficiency when handling different levels of dangers.
    \item Emergencies typically encountered by multi-UGV and multi-UAV systems are summarized and modeled into three scenarios: hidden obstacle, turbulence, and strong wind. The simulation results reveal that with a small sacrifice of efficiency, CCR significantly reduces the risky robotic behavior and largely improves the MRS resilience towards common unexpected dangers. 
\end{enumerate}

\section{RELATED WORK}

\textit{Safe Multi-Robot System.} Due to its attractive capability in performing large-scale and complex tasks, the safety of robot teams has been researched in recent years. Some methods~\cite{fung2019coordinating, desai2019efficient, freda20193d} used a prior map to help avoid unexpected or dynamic obstacles. One of the popular ways to achieve safety in the MRS was eliminating dangerous behaviors among robots and between robots and the environment \cite{glotfelter2017nonsmooth, wang2017safety}. Such algorithms, like control
barrier function \cite{ames2019control, taylor2020learning}, formulated the problem into a constrained optimization problem and  planned collision-free trajectories.  A more relaxed approach tolerated unsafe behaviors and robot failures \cite{portugal2013scalable,park2017fault, park2016efficient}, which was more robust to complex environments. However, both methods are based on an explicit mathematical model of the emergency situation, which assumes full knowledge of its type, size, location, and intensity. On the other hand, the CCR algorithm introduced in this paper can adapt to emergency situations with high randomness and poor observability. 

\textit{Safe Reinforcement Learning for MRS.} There has been an increasing effort to incorporate reinforcement learning (RL) into safe multi-robot systems.  A carefully-designed reward function was usually a necessity for such methods to work. Existing related works were classified into two paradigms: communicative emergency reflex and non-communicative emergency reflex. In \cite{long2018towards}, a multi-scenario multi-stage training framework was developed to train a fully decentralized sensor-level collision avoidance policy for MRS. To improve system coordination without adding more communication assumptions, Chen et al. proposed to use value networks to directly generate trajectories, which showed great performance \cite{chen2017decentralized} for non-communicating MRS. Although these previous works developed safer robot behaviors, important real-world metrics such as minimum safety distance and group response time were not considered. 
In contrast, CCR offers a tunable parameter that controls these properties through controlling the trade-off between efficiency and safety. This allows for more leniency when applying to real-world systems.

\section{Collective  Conditioned  Reflex (CCR)}

Ensuring safety and efficiency during emergency reactions in multi-robot systems has been an open research topic. Existing approaches tend to focus only on control commands at the physical level while ignoring heuristic awareness of the danger that the robot itself can build. To help build this recognition and use it to achieve better performance, we propose Collective Conditioned Reflex (CCR), a novel bio-inspired emergency reaction mechanism that can be attached to most MARL algorithms. 

\textit{Preliminaries of MRS Reinforcement Learning. }The standard setup of reinforcement learning assumes a Markov decision process (MDP) model for the agent-environment interaction. A MDP consists of a 5-tuple $(\mathcal{S}, \mathcal{A}, \mathcal{R}, \mathcal{T}, \gamma)$, representing the set of states, set of actions, reward function, transition function, and the discount factor, respectively. The agent is tasked with learning a policy $\pi:\mathcal{S}\times \mathcal{A}\to [0,1]$ such that the total expected discounted reward $\sum_{t}\gamma^tr^t$ is maximized. This study uses the multi-agent extension of the MDP, where each of the $N$ agents has its own state $s_1,\dots, s_N$ and action $a_1,\dots,a_N$. The agents are tasked with maximizing their own expected total discounted reward.

\subsection{Overview of CCR Design}
\begin{figure}[t]
\flushleft
\includegraphics[width=1\linewidth]{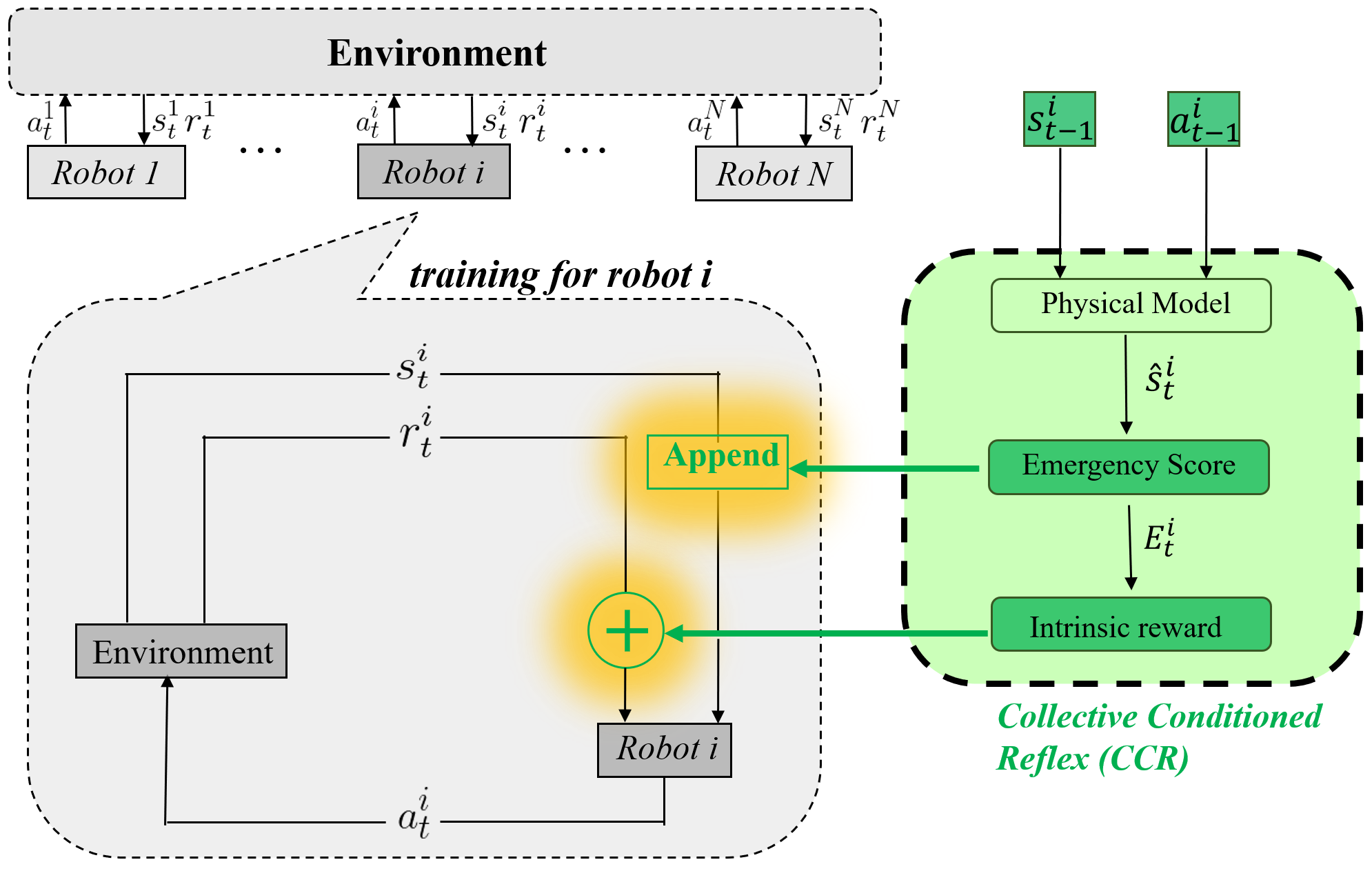}
\caption{The overall framework of Collective Conditioned Reflex (CCR)}
\label{fig:overall framework}
\vspace{-0.6cm}
\end{figure}

Fig. \ref{fig:overall framework} is an overview of the CCR framework. On the left-hand side, a MARL algorithm learns a policy for each of the $N$ robots by interacting with the environment. Let $s_t^i$, $r_t^i$, and $a_t^i$ denote the state, the reward, and the action of $i$th robot at time $t$, respectively. The CCR module comes into play by redefining the reward and state received by the robots based on the emergency score derived from the learned physical model. The physical model is a group of polynomial dynamics functions with learnable parameters. Its input is robots' control commands while the output is the robots' next states. The previous step state $s_{t-1}^i$ and action $a_{t-1}^i$ are stored and used to compute the ideal current state $\hat s_t^i$ for each robot. This represents how the environment dynamics is expected to flow without disturbance. Then, the difference between $\hat s_t^i$ and $s_{t}^i$ is used to compute an emergency score for each robot, which is appended to  other robots' states within robot $i$'s communication range.  This gives the robots more explicit information about the emergency. Then, the emergency scores are used to compute an intrinsic reward for each robot via a heuristic algorithm, which is added to the reward received from the environment. The intrinsic reward provides the robots with additional information about the potential consequences of actions. Robots will be discouraged from getting close with other robots in danger and encouraged to approach other robots in safety.

\subsection{Emergency Score Assessment}

The idea of emergency score derives from the perception of animals to external emergencies and is used to measure how dangerous they are. In general, animal groups have different levels of responses to crises of different emergency scores. In MRS, emergency scores can remind robots to learn adaptive local reaction policies to different levels of danger. CCR assumes access to the previous step transition of robot $i$ at time $t-1$: $(s^i_{t-1}, a^i_{t-1}, r^i_{t-1}, s^i_{t})$. The predicted states $\hat s_t^i$ is obtained by passing the last step transition to the physical model: 
$$\hat s_t^i\sim \mathcal{M}(s_{t-1}^i,a_{t-1}^i)$$
where the states $s_{t-1}^i$ include the following information: the Cartesian coordinates of robot $i$, the velocity of robot $i$, the distances and directions to other robots, the distance and direction to the destination. The emergency score is computed as a form of prediction error and then propagated to other nearby robots within its communication range:
$$E^i_t:=\|\hat s_t^i-s_t^i\|^2$$
In practice, we observe that smoothing the emergency score with a rolling average can further strengthen the performance. Let $m$ be the number of time steps to take on average. Then,
$$E^i_t\gets \frac{1}{m+1}\sum_{k=0}^{m}E^i_{t-k}$$
This step is repeated for all $N$ robots.

\subsection{Potential Consequences Evaluation}
Reward in reinforcement learning can be understood as an evaluation of how advantageous the action of a robot is. In the traditional setting, this evaluation is done by the environment, and the robot gets the reward as a return. However, it is also possible for the robot to evaluate with a metric different than the environment. This is the motivation behind intrinsic reward. In this work, the intrinsic reward is designed to estimate the potential consequences caused by the emergency.

Specifically, the emergency score introduced in the previous section is used to compute the intrinsic reward as the estimation of potential consequences. The intrinsic reward is then augmented to the reward returned by the environment to form the final reward used to train policy for each robot.

First, the emergency scores for robots are compared pairwise. Intuitively, within each pair, the robot with the lower emergency score should be at a safer state than the other robot. When this difference is significant, it hints the existence of an ``emergency'' at the other robot's location. For example, when robot $i$ is at a safer state than robot $j$, $E_i<E_j$ and therefore robot $i$ should be rewarded for moving away from $j$ and penalized for moving towards $j$. This motivates us to define the \textit{change in distance} $h_i(j)$, which corresponds to the change in Euclidean distance between robot $i$ and robot $j$ from time $t$ to $t+1$,
$$h_i(j)=\|s_{t+1}^i-s_{t}^j\|_2-\|s_{t}^i-s_{t}^j\|_2$$ 
Then, we define the intrinsic reward for robot $i$ induced by robot $j$,
$$R_{in}^i(j)=\lambda\left((E_j-E_i)\frac{h_i(j)}{1+\|s_{t}^i-s_{t}^j\|}\right)$$
In the previous example, $R^i_{in}(j)$ is negative when robot $i$ is moving towards robot $j$, and positive when robot $i$ is moving away from robot $j$. The opposite is true when robot $j$ is at a safer state than robot $j$, which means $E_j<E_i$. Note that robot $i$ computes its intrinsic reward entirely through observing the effect of the emergency on robot $j$, assuming no knowledge on the emergency itself. Summing the intrinsic reward induced by other robots will give us the final intrinsic reward for robot $i$:
$$R_{in}^i=\sum_j R_{in}^i(j)$$

Each $r_t^i$ received from the environment is added with the intrinsic reward, with a tunable $\lambda$ coefficient multiplying $R_{in}^i$. The augmented reward replaces the reward obtained from the environment, and the transition is then saved to the experience replay or summed into the return. Regular MARL algorithms take on from here.

\subsection{Developing Emergent Collective Behaviors}
Emergency scores prompt robots with immediate emergency hazard information which will stimulate them to take the corresponding reactions at the same time. Otherwise, robots will miss the best first reaction time window that may lead to global failure. Since the estimation of potential consequences is computed from the emergency scores of other robots, the behavior of each robot is influenced heavily by the states and actions of its neighbors. Thus, CCR enhances emergent behavior in the MRS by actively propagating the emergency information from local to the whole team. Local robots in the danger zone will be self-organized to form local first response to the emergency.

\section{Evaluation}
\subsection{Simulation Setting}
A simulated environment is constructed to evaluate the effectiveness of the CCR algorithm. Similar to other related researches~\cite{liu2019trust, luo2020multi, li2017decentralized}, the goal of the simulations is to investigate how the behaviors of the robots change when CCR is applied and to quantitatively measure the effect of this change. Our simulation environment is based on the Multi-Agent Particle Environment\footnote{https://github.com/openai/multiagent-particle-envs} \cite{lowe2017multi}, a simple planar multi-agent particle world based on basic simulated physics with velocity as control inputs. The environment is modified to include more objects and improve rendering. We implement three scenarios to simulate how MRS might encounter dangers in the real world: \textit{turbulence, strong wind}, and \textit{hidden obstacle}. Specifically, \textit{turbulence} and \textit{strong wind} aim at simulating multi-UAV systems under the influence of unpredictable turbulent or laminar airflow, while \textit{hidden obstacle} aims at simulating multi-UGV systems encountering obstacles with low detectability or in adversarial weather. The simulation environment, although basic, serves as a proof of concept for CCR's overall framework and fundamental idea. Details of the simulation setting and scenarios are shown in Fig. \ref{fig:emergencies} and introduced as follows:

\begin{figure}[t]
\centering
\includegraphics[width=0.44\textwidth]{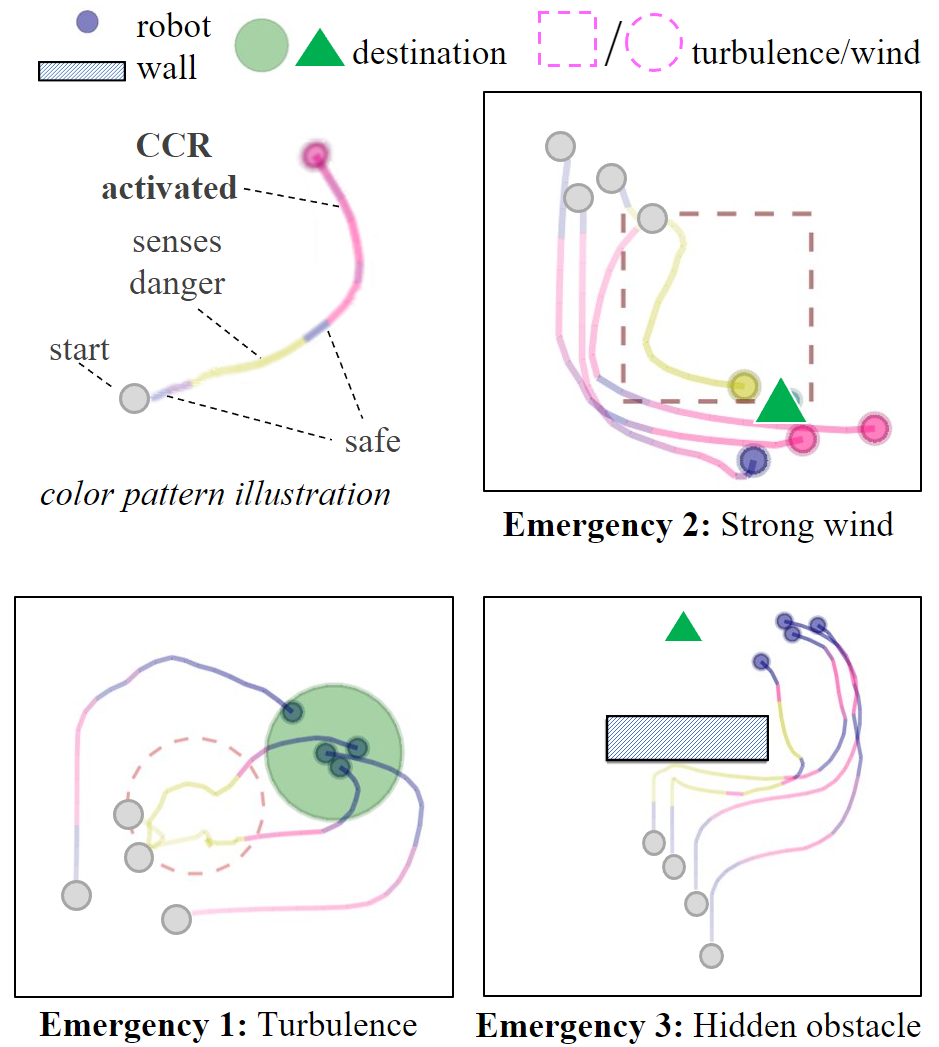}
\caption{The initial and end states of the three scenarios. ``Unsafe behavior'' is defined as entering the turbulence/wind area or colliding with the obstacle. The color pattern used to describe the robot trajectory is explained on the top left.}
\label{fig:emergencies}
\vspace{-0.6cm}
\end{figure}

\textit{Emergency 1: turbulence:} The robots are tasked to find their way to a circular target area marked with green color. A circular turbulence area will appear on their way and apply forces with random directions on them. Coordinates of the turbulence area is only observable for robots inside the area. The task is considered successful for each robot that arrives in the target area.

\textit{Emergency 2: strong wind:} The robots are tasked to navigate to a green target point. A square strong wind area with random sizes will appear on their way and apply force with a fixed direction. Coordinates of the strong wind area is only observable for robots inside the area. The task is considered successful for each robot that arrives at the other side of the strong wind area.

\textit{Emergency 3: hidden obstacle:} The robots are tasked to navigate to a green target point. A rectangular solid obstacle with random lengths is located on their way. Robots can only sense whether they are in collision with the obstacle, but not the coordinates of the obstacle. The task is considered successful for each robot that arrives at the other side of the obstacle.

In all scenarios, the robots are homogeneous and have a fixed communication range. Any robot can send and receive simple information such as a floating point number to other robots only if they are within its communication range. The environmental rewards in such three scenarios are all based on the distance from each robot to the destination. 

\textbf{Metrics.}
We evaluate the performance of MARL algorithms equipped with the CCR module with four metrics designed to quantify both the safety and the efficiency of the robot trajectories. The safety aspect is evaluated with the average distance from each robot to the center of the danger area at every time step and the proportion of time steps robots are in dangerous states in the whole episode (dangerous behavior frequency). The efficiency aspect is evaluated with the proportion of robots that successfully accomplish the task (success rate) and the reward. 

\textbf{Baseline.} 
Simulations are conducted with three popular MARL algorithms -- independent DDPG \cite{lillicrap2015continuous}, MADDPG \cite{lowe2017multi}, and MAAC \cite{iqbal2019actor} as baselines. In addition, we also implement a common safety mechanism of augmenting the external reward with a penalty term when a dangerous step is taken, named as pessimistic DDPG, pessimistic MADDPG, and pessimistic MAAC, respectively \cite{wu2022improved, zhu2020multi}. We take them as baselines to help demonstrate the effectiveness of CCR in improving MRS safety. The state space is continuous for all three algorithms, but the action space is continuous only for IDDPG and MADDPG and discrete for MAAC. This is to diversify the settings that we test the CCR module on and contribute to proving the broad applicability of our approach. 

To illustrate that robots equipped with CCR can have a timely reaction effect, different colors are assigned to the trajectory of robots when CCR is running in different steps. The color pattern is explained in Fig. \ref{fig:emergencies}. The trajectory is blue when the robot is operating normally and the CCR module is on standby, yellow when the robot senses the danger, and pink when CCR is activated and triggers an intrinsic reward. Fig. \ref{fig:emergencies} also illustrates the unsafe behavior and success criteria as described above.

The robot operator can tune the efficiency-safety tradeoff by changing the hyperparameter $\lambda$, which controls the weight of the intrinsic reward added to the environment reward. After conducting a parameter search, we find that setting $\lambda=30$ gives a fairly balanced trade-off between efficiency decrease and safety improvement, and generalizes well across all three scenarios. Therefore, all simulations shown here are conducted with $\lambda=30$. MADDPG and DDPG policies are trained for $16000$ steps, and the MAAC policy is trained for $16000\times 35$ steps. 

\begin{figure}[t]
\centering
\includegraphics[width=0.47\textwidth]{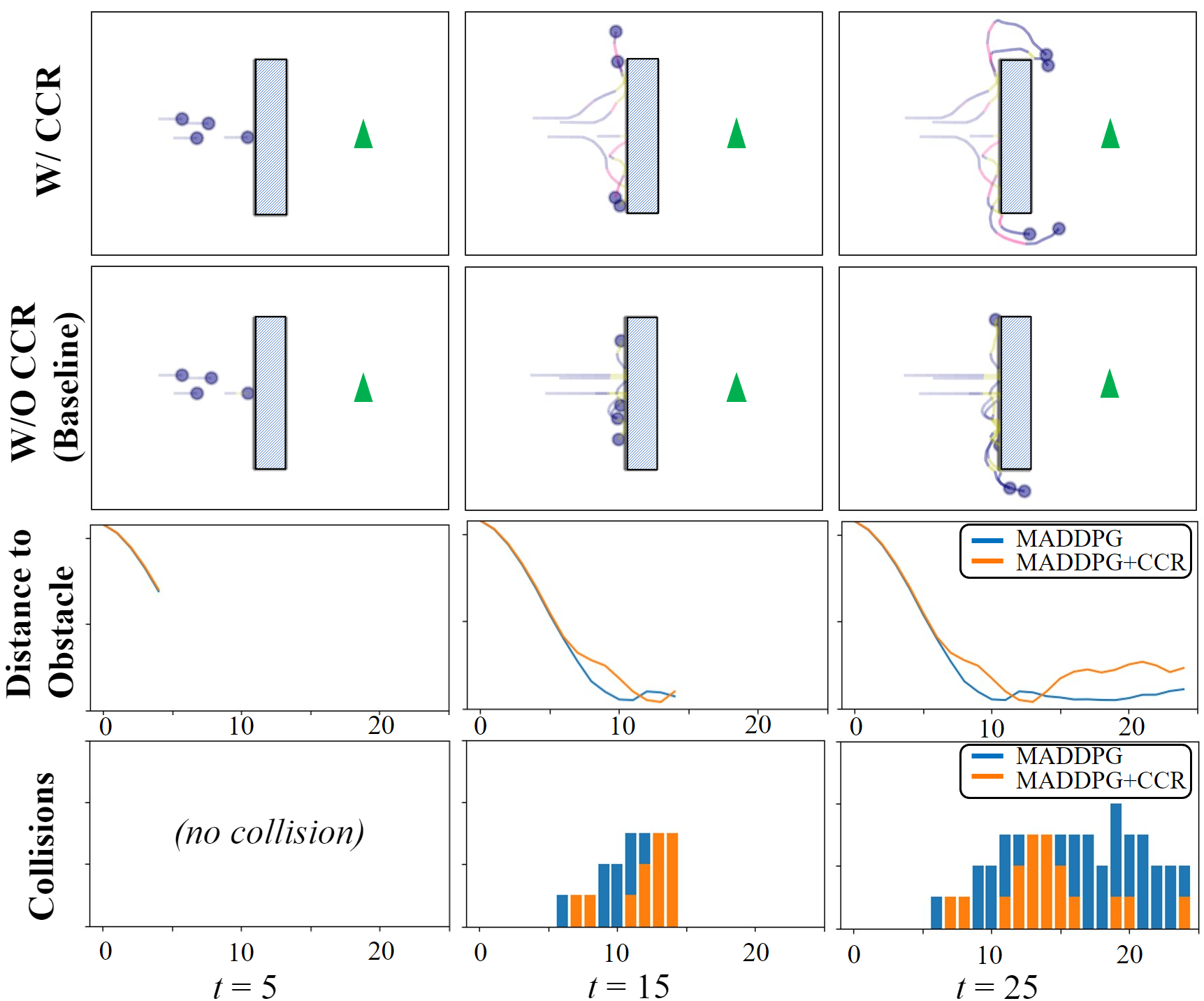}
\caption{The trajectory comparison between MADDPG with and without CCR module in hidden obstacle scenario.} 
\label{fig:wall_traj}
\vspace{-0.3cm}
\end{figure}

\subsection{Overall Performance}
We measure the task success rate and the dangerous behavior frequency to validate CCR's overall effectiveness in safety improvement and to investigate the tradeoff between safety and efficiency. The results are shown in Tables \ref{t1} and \ref{t2}. Compared with policies trained by the original MARL algorithms, it is found that with exception of \textit{strong wind} with MAAC, the dangerous behavior frequency of those trained by CCR-aided algorithms is lower in all scenarios. On average, applying CCR can effectively reduce the dangerous behavior frequency by $42.24\%$. The effect on task success rates vary across different scenarios. For \textit{hidden obstacle}, the success rate of policies trained by MARL+CCR is significantly higher than those of the original algorithms. For \textit{turbulence}, however, MARL+CCR algorithms suffer from an average of $23.81\%$ lower success rate. For \textit{strong wind}, the success rate of MARL+CCR is higher than the original when using DDPG and MAAC, but lower when using MADDPG. This discrepancy is caused by the different dynamics between robots and the emergency in the three scenarios. For example, in \textit{hidden obstacle}, MARL policies are often held back by the obstacle, which simultaneously increases chance of collision and decreases the task success rate. MARL+CCR policies can more effectively circumvent the obstacle, which decreases chance of collision and increases the task success rate. On the other hand, the turbulence area in \textit{turbulence} can be traveled through, which leads to both higher success rate and higher dangerous behavior frequency for MARL policies. MARL+CCR policies often try to avoid the turbulence area, which contributes to decreased dangerous behavior frequency but also decreased success rate. 

Compared with policies trained by the modified pessimistic MARL algorithms, those trained by CCR-aided algorithms are safer  in all scenarios. On average, the dangerous behavior frequency of MARL+CCR policies is $38.37\%$ lower than MARL (pessimistic) policies. The task success rate for MARL+CCR algorithms is higher than that of MARL (pessimistic) algorithms in 7 out of 9 cases, with the exception being MADDPG in \textit{strong wind} and MAAC in \textit{turbulence}. The success rate is increased by $35.90\%$ on average. The pessimistic approach cannot achieve ideal performance because information about the emergency events is sparse and myopic in nature: robots can only sense collision with the hidden obstacle but not its exact location and size, and they can only observe the turbulence or wind area when they are already inside these areas. This lack of information removes assumptions about prior knowledge to the environment, making MARL algorithms more practical for real world MRS applications. Because MARL (pessimistic) algorithms cannot associate the pessimistic penalty with an apparent change in the observations, they will either fail to improve safety or overfit to an overly cautious policy and reduce task success rate.

\begin{figure}[t]
\centering
\includegraphics[width=0.50\textwidth]{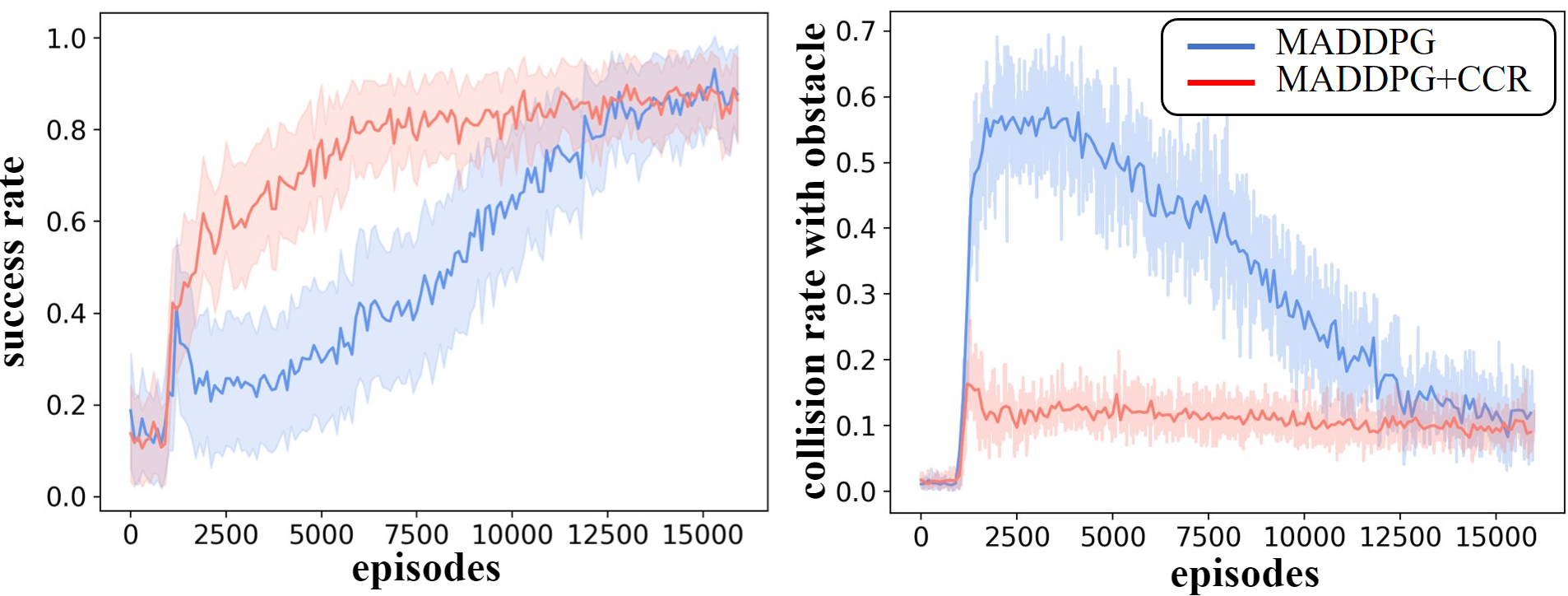}
\caption{Number of collisions with the obstacle and success rate in the hidden obstacle scenario.} 
\label{suc_col}
\vspace{-0.7cm}
\end{figure}

To examine the effect of CCR during training, the time-driven plot of dangerous behavior frequency and task success rate during training for both original MADDPG and MADDPG+CCR are shown in Fig. \ref{suc_col}. It is found that CCR facilitates safe exploration in the training process, with an episodic dangerous behavior frequency of mostly below 0.20, compared to the maximum of 0.56 for the original MADDPG. Meanwhile, the success rate of MADDPG+CCR remains higher or on par with that of the original MADDPG throughout training. During the training process, the cumulative number of dangerous behaviors of MADDPG+CCR is only $28.09\%$ of that of MADDPG. We examine the results from other scenarios and algorithms and observed similar trends. These results show that CCR is capable of reducing training time dangerous behavior frequency in a wide range of MRS tasks. This offers great potential for deploying the algorithm to real-world systems, where exploration in complex environments can be costly and dangerous.

Meanwhile, we also notice some limitations of our approach. From Fig. \ref{suc_col}, we can observe that CCR may slow down the convergence rate compared with baselines, which can be attributed to the appended states (higher dimensions) and intrinsic rewards being more difficult to infer. Another disadvantage is that CCR needs to trade efficiency for safety to some extent.

\begin{table}[t]
\caption{Dangerous behavior frequency in different scenarios}
\begin{tabular}{ m {3 cm}  m {1.2 cm}  m {1.2 cm}  m {1.2 cm} }
\label{t1}
 & hidden obstacle & turbulence & strong wind \\
\hline
 DDPG &             40.86\% & 48.03\% & 32.63\% \\ 
 DDPG (Pessimistic) &22.80\% & 54.70\% & 19.97\%\\
 DDPG+CCR &         \textbf{8.41\%} & \textbf{23.35\%} & \textbf{19.24\%} \\  
\hline
 MADDPG &               40.92\% & 52.35\% & 19.22\% \\
 MADDPG (Pessimistic) & 22.42\%& 57.86\% &21.96\%\\
 MADDPG+CCR &           \textbf{15.27\%} & \textbf{34.97\%} & \textbf{17.97\%}  \\
\hline
 MAAC &                 69.34\% & 52.25\% & \textbf{12.15\%}  \\
 MAAC (Pessimistic) &   42.87\% & 25.74\% & 25.73\% \\
 MAAC+CCR &             \textbf{28.89\%} & \textbf{9.73\%} & 16.25\%  \\
\hline
\end{tabular}
\vspace{-0.3cm}
\end{table}

\begin{table}[t!]
\caption{Success rate in different scenarios}
\begin{tabular}{ m {3 cm}  m {1.2 cm}  m {1.2 cm}  m {1.2 cm} }
\label{t2}
 & hidden obstacle & turbulence & strong wind \\
\hline
 DDPG &             36.22\% & 97.48\% & 73.44\% \\ 
 DDPG (Pessimistic) &73.16\% & 56.18\% & 72.45\%\\
 DDPG+CCR &         88.69\% & 70.34\% & 81.74\% \\  
\hline
 MADDPG &               33.80\% & 96.62\% & 70.40\%  \\
 MADDPG (Pessimistic) & 73.53\% & 54.77\% & 79.75\% \\
 MADDPG+CCR &           75.79\% & 84.00\% & 77.15\%  \\
\hline
 MAAC &                 1.71\% & 97.19\% & 84.26\%  \\
 MAAC (Pessimistic) &   8.43\% & 78.41\% & 78.45\%  \\
 MAAC+CCR &             26.71\% & 67.50\% & 84.51\%  \\
\hline
\end{tabular}
\vspace{-0.4cm}
\end{table}

\subsection{Behavioral Analysis}
Fig. \ref{fig:wall_traj} illustrates how the behavior of the robots in the hidden obstacle scenario is substantially affected when the CCR module is introduced into training. In the hidden obstacle scenario, since the robots can only sense the obstacle when they get sufficiently close to it, all the robots trained with the baseline MADDPG algorithm will repeat the process of running into the obstacle, then moving away from it. On the other hand, for the robots trained with CCR-aided MADDPG, only the first two robots will run into the obstacle, while others steer away from it without even observing the obstacle. This is because the obstacle is not present in the physical model, and therefore robots that run into the obstacle will have high emergency scores because they do not expect there to be an obstacle. Other robots will thus receive positive intrinsic rewards for moving away from them and negative intrinsic rewards for moving towards them. Essentially, the following robots will obtain danger information from observing the emergency reaction behavior of the leading robot and react in advance. This is eventually reflected in the policy after training. The behavior is confirmed by the average distance to the obstacle in Fig. \ref{fig:wall_traj}, which shows that in total, the robots trained with CCR-aided MADDPG get much farther to the obstacle than those trained with MADDPG, by approximately 1.2x to 4.7x. We also provide a video detailing this phenomenon\footnote{https://youtu.be/AswbaBMJi9E}. 


\section{Conclusion}

\addtolength{\textheight}{-1cm}   



This paper introduces Collective Conditioned Reflex (CCR), a biology-inspired fast emergency reaction module based on multi-agent
reinforcement learning for multi-robot systems.
In the future, more realistic 3D physical dynamic simulators can be deployed to evaluate the effectiveness of our method. 

\bibliographystyle{unsrt}
\bibliography{ref}
\end{document}